\documentclass[10pt,conference]{IEEEtran}
\usepackage[utf8]{inputenc}

\usepackage{times}  
\usepackage{helvet}
\usepackage{courier}  
\usepackage{graphicx} 
\usepackage{amsmath}
\usepackage{siunitx}
\usepackage{amsthm}
\usepackage{relsize}
\usepackage{amssymb}
\usepackage{algorithm}
\usepackage{booktabs}
\usepackage{verbatim} 
\usepackage{balance}
\usepackage{graphicx}
\usepackage{tikz}
\usepackage{algpseudocode}
\usepackage[bottom]{footmisc}
\usetikzlibrary{fit,positioning}
\usepackage[utf8]{inputenc}
\usepackage[english]{babel}
\usepackage{caption}
\usepackage{mathtools}
\usepackage[hyphens]{url}

\usepackage{xcolor}
\usepackage{float}

\usepackage{subcaption}

\usepackage{fancyhdr}
\usepackage{kantlipsum}
\fancypagestyle{plain}{
\fancyhf{}
\fancyhead[C]{Conference on \LaTeX} 
} \usepackage{eso-pic}

\IEEEoverridecommandlockouts
\begin{document}

\title{A Deep Reinforcement Learning Approach towards Pendulum Swing-up Problem based on TF-Agents}

\author{\IEEEauthorblockN{Yifei Bi, Xinyi Chen, Caihui Xiao}
\IEEEauthorblockA{Department of Statistics,
Columbia University in the City of New York, New York, USA\\
\{yb2456, xc2464, cx2225\}@columbia.edu}
}

\maketitle

\begin{abstract}
Adapting the idea of training CartPole with Deep Q-learning agent, we are able to find a promising result that prevent the pole from falling down. The capacity of reinforcement learning (RL) to learn from the interaction between the environment and agent provides an optimal control strategy. In this paper, we aim to solve the classic pendulum swing-up problem that making the learned pendulum to be in upright position and balanced. Deep Deterministic Policy Gradient algorithm is introduced to operate over continuous action domain in this problem. Salient results of optimal pendulum are proved with increasing average return, decreasing loss, and live video in the code part.
\end{abstract}
\section{Introduction}
\label{sec:intro}
Alongside supervised learning and unsupervised learning, reinforcement learning (RL) is one of three machine learning paradigms, which is concerned with how software agents ought to take actions in an environment in order to maximize some notion of cumulative reward. Unlike supervised learning that learning a relationship between i.i.d input-output pairs and using data being representative of all possible scenarios, RL uses learned knowledge from the environment to provide optimal controllers and such controllers can adapt to a changing environment, and its training data cannot capture all possible scenarios. Finding a balance between exploration and exploitation is of importance.

In order to excel the Pendulum game in OpenAI Gym, we choose and implement suitable agent imported from a deep RL library called TF-agents \ref{2.2}. The pendulum is a classical benchmark problem for the purpose of continuous control, which introduces in detail in subsection \ref{2.1}. With the help of tf-agents, now we are able to efficiently implement more complicated agent, like the Deep Deterministic Policy Gradient (DDPG) which is used in our project. DDPG is an algorithm which concurrently learns a Q-function and a policy to solve the problem. This algorithm is presenting in the subsection \ref{2.3}. The performance of the controller and experiment process like parameter tuning are in the section \ref{3}. Conclusion and discussion for further steps are in the section \ref{4}.
\section{Solution}
\subsection{Environment}
\label{2.1}

The inverted pendulum swingup problem is a classic problem in the control literature. In this version of the problem, as introduced in \cite{pendulum-v0}, the pendulum starts in a random position, and the goal is to swing it up so it stays upright.

\begin{figure}[H]
    \centering
    \includegraphics[scale=0.5]{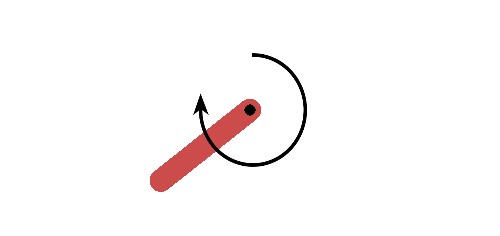}
    \caption{Pendulum-v0}
    \label{fig:1}
\end{figure}

\begin{table*}[h]
\caption{Environment Spaces}
\label{tab:1} 
\centering
\resizebox{2\columnwidth}{!}{%
\begin{subtable}{0.3\linewidth}
\caption{State Space (Continuous)}
\label{tab:1a}
\centering
\begin{tabular}{|c|c|c|c|}
     \hline
     \textbf{Num} & \textbf{State} & \textbf{Min} & \textbf{Max}\\
     \hline
     0 & cos$(\theta)$ & -1.0 & 1.0\\
     1 & sin$(\theta)$ & -1.0 & 1.0\\
     2 & $\theta$ dot & -8 & 8\\
     \hline
\end{tabular}
\end{subtable}%
\begin{subtable}{0.3\linewidth}
\caption{Action Space (Continuous)}
\centering
\label{tab:1b}
\begin{tabular}{|c|c|c|c|} 
     \hline
     \textbf{Num} & \textbf{Action} & \textbf{Min} & \textbf{Max}\\ 
     \hline
     0 & Joint Effort & -2.0 & 2.0\\
     \hline

\end{tabular}
\end{subtable}%
}
\end{table*}

Table \ref{tab:1} above shows the state space and action space of pendulum environment where $\theta$ represents the pendulum angle, which start randomly from $-\pi$ to $\pi$. The default reward function depends on $\theta$. To get maximum reward, the angle needs to remain at zero, with the least rotational velocity, and the least effort.

\subsection{TF-Agents}
\label{2.2}
In this project, we trained reinforcement-learning agent on the Pendulum environment using the TF-Agents library. TF-Agents is a robust, scalable and easy to use RL Library for TensorFlow, which is compatible with other TF high-level API's and therefore is extensively resourceful. It standardizes common steps of RL and thus helps researchers to try and test new RL algorithms quickly. Moreover, it is also well tested and easy to configure with gin-config \cite{TF-AgentTutorial}.

\subsection{Deep Deterministic Policy Gradient (DDPG)}
\label{2.3}
In Advanced Machine Learning class, we mainly learned the great Q-Learning algorithm and Deep Q-Networks (DQN), which arouse our interest and laid the foundation for our further study in RL area. However, while DQN solves problems with high-dimensional observation spaces, it can only handle discrete and low-dimensional action spaces. Obviously, as introduced in subsection \ref{2.1}, the pendulum environment has continuous action space. Thus, we studies many other methods and finally chose DDPG agent, which is optimal in our case. Before getting to the DDPG algorithm, there are other two RL methods need to be understood besides DQN: Policy Gradient (PG) and Actor-Critic. 

PG is widely suited in RL problems with continuous action spaces. The basic idea is to represent the policy by a parametric probability distribution $\pi_\theta(a|s)=P[a|s;\theta]$ that stochastically selects action $a$ in state $s$ according to parameter vector $\theta$. PG algorithms proceed by sampling this stochastic policy and optimizing the policy parameters in the direction of greater cumulative reward through gradient ascent \cite{silver2014deterministic}. 

The Actor-Critic algorithm represents the policy function and value function independently. Policy function works as an Actor which produce an action based on current status while value function is a Critic that evaluates the Actor's action and produces TD error signal to determine how the action should change. 

Combining methods above, DDPG is a policy gradient algorithm that uses a stochastic behavior policy for good exploration but estimates a deterministic target policy, which is much easier and more efficient to learn on continuous action space \cite{DDPG-TF}. DDPG learns directly from state spaces follow the policy gradient to maximize performance in contrast with DQN that learns indirectly through Q-table. It also employs Actor-Critic model with some of the deep learning tricks: a replay buffer, online and target Q networks for Critic, and deterministic online and target policy networks for Actor. However, rather than directly copying weights from online to target, DDPG uses soft target updates to improve stability. In addition, Actor directly maps states to actions instead of outputting the probability distribution. Replay Buffer allows the agent to learn offline by gathering experiences collected from environment and sampling a random batch of experience from the Buffer which are uncorrelated.

The parameters of four networks in DDPG are introduced in Table \ref{tab:parameters}.

\begin{table*}[h]
\caption{Parameters of four networks in DDPG}
\label{tab:parameters}
\centering
\begin{tabular}{c c c c}
    \toprule
     $\theta^\mu$:  & deterministic online policy function $\mu$ & $\theta^{\mu'}$: & target policy function $\mu'$\\
     $\theta^Q$: & online Q network $Q$ & $\theta^{Q'}$: & target Q network $Q'$\\
     \bottomrule
\end{tabular}
\end{table*}


\begin{figure*}
  \centering
  \includegraphics[width=0.95\textwidth]{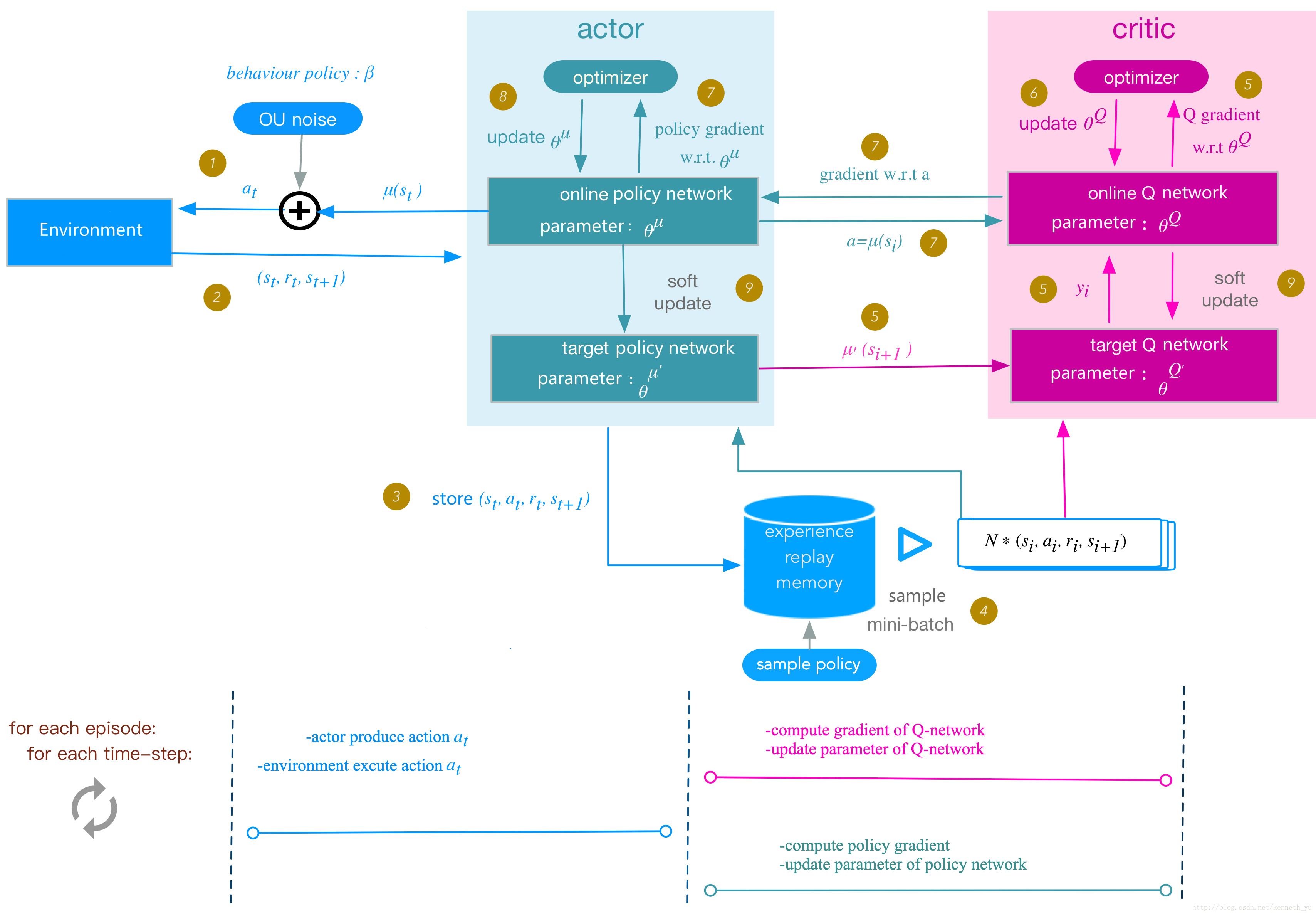}
  \caption{Workflow of DDPG Algorithm}
  \label{fig:2}
\end{figure*}

Figure \ref{fig:2} shows the entire workflow of DDPG:
\begin{itemize}
    \item Step 0 (Initialization): Randomly initialize $\theta^\mu$ and $\theta^Q$, and copy to target networks: $\theta^{\mu'} \leftarrow \theta^\mu$, $\theta^{Q'} \leftarrow \theta^Q$. Initialize Replay Buffer.
    \item Step 1 (For loops): Run the following for loops (an outer for loop and an inner for loop):
    \begin{itemize}
        \item[--] \textit{For each episode:} Initialize a Uhlenbeck-Ornstein random process $\mathcal{N}$ for action exploration
            \begin{itemize}
            \item \textit{For each time step t:}
                \begin{enumerate}
                \item  Select action $a_t = \mu(s_t|\theta^mu)+\mathcal{N}_t$ where $\mu$ is the current online policy and $\mathcal{N}$ is the exploration noise.
                \item Execute $a_t$, get reward $r_t$ and next state $s_{t+1}$.
                \item  Store this transition $(s_t, a_t, r_t, s_{t+1})$ to Replay Buffer.
                \item  Sample a random minibatch(N) of transitions $(s_i, a_i, r_i, s_{i+1})$ from Replay Buffer.
                \item  $y_i = r_i+\gamma Q'(s_{i+1},\mu'(s_{i+1}|\theta^{\mu'}))$
                \item  Update online Q-network by minimizing loss function(MAE was used here) $L=\frac{1}{N}\sum_i|y_i-Q(s_i,a_i|\theta^Q)|$ with Adam optimizer.
                \item  Update online policy network using policy gradient
                $\nabla_{\theta^\mu}J \approx \frac{1}{N}\sum_i \nabla_a Q(s,a|\theta^Q)|_{s=s_i,a=\mu(s_i)} \nabla_{\theta^\mu}(\mu(s_i|\theta_\mu))|_{s_i}$ with Adam optimizer.
                \item Soft update target networks $\mu'$ and $Q'$:
                $$\theta^{\mu'} \leftarrow \tau\theta^\mu +(1-\tau)\theta^{\mu'};$$ $$\theta^{Q'} \leftarrow \tau\theta^Q + (1-\tau)\theta^{Q'}.$$
                \end{enumerate}
            \item \textit{end for time step}
            \end{itemize}
       \item[--] \textit{end for episode} 
    \end{itemize}
\end{itemize}

\subsection{Experiment Details}
\label{2.4}
Adam optimizer is applied to both actor and critic networks with a learning rate of $10^{-4}$ and $10^{-3}$ respectively. We used a discount factor of $\gamma = 0.99$, the soft target updates of $\tau = 0.05$ and the soft target frequency of period $C = 5$. The networks had 2 hidden layers with 400 and 300 units respectively. We trained with mini batch sizes of 32 and used a replay buffer size of 100,000. Training loss is printed every 20 iterations and average return is printed every 1,000 iterations. Collecting steps per iteration is 1. For the exploration noise process we used temporally correlated noise in order to explore well in physical environments that have momentum. We used an Ornstein Uhlenbeck process (Uhlenbeck $\&$ Ornstein, 1930) with $\theta = 0.15$ and $\sigma = 0.3$. The Ornstein Uhlenbeck process models the velocity of a Brownian particle with friction, which results in temporally correlated values centered around 0 \cite{lillicrap2015continuous}. In section \ref{3}, we will discuss the results from tuning parameters.
\section{Evaluation} \label{3}

We oerates 100,000 number of iterations to train the inverted pendulum swing-up problem. Pendulum-v0 is still a learning environment, which does not have a specified reward threshold at which it's considered solved. In our case, average return increased from below -1000 to about -140 and then becomes converged. This training process took 29 mins to run on Google Colab. Meanwhile, Pendulum was played several times, the live video showed it quickly swung up and could stay upright.

\begin{figure}[H]
    \centering
    \includegraphics[scale=0.25]{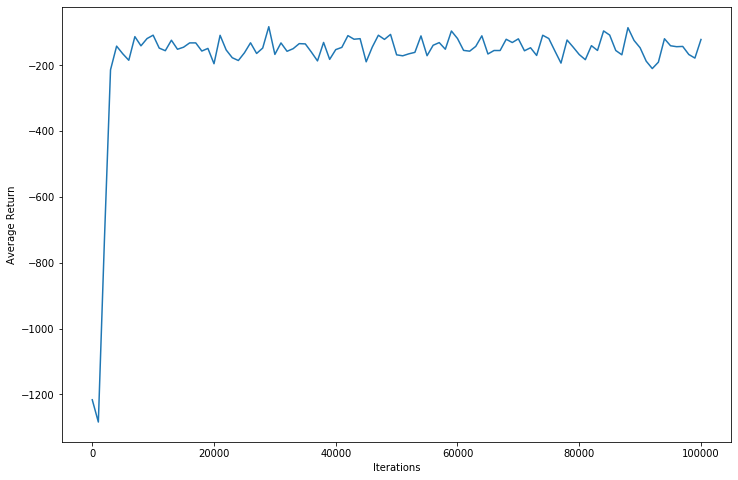}
    \caption{Average Return}
    \label{fig:3}
\end{figure}

During training process, we trains the system several times with different number of iterations and parameters. 20,000, 50,000 and 1,000,000 iterations are all conducted for training the system. As we notice in Figure \ref{fig:3}, average return converges at early stage. Video shows the pendulum is improved with 20,000 iterations but is still not at an optimal upright position. We increases our training iterations to 50,000, then AR increases as well and performance becomes better. Lastly, we try 1,000,000 iterations, and the overall performance achieves optimum. Using DDPG agent, our problem can be solved within 100,000 iterations. For other complicated problems like Atari domain, DDPG is able to resolve them in fewer steps than Deep Q-learning. In terms of batch size, we tested 32, 64 and 100 respectively. Size of 32 is able to solve our problem better. More complex tasks tend to have noisier gradients and increasingly large batch sizes are likely to become useful.

\setlength{\belowcaptionskip}{-10pt}
Besides Average Return, Training Loss is also used as a measure of performance of our agent. As defined in agent, loss output is the total loss of actor and critic emerging during learning process. At first, the most commonly use loss functions Huber Loss and Mean Squared Error(MSE) are chosen and tested. As building in the package, loss functions only change the way of computing critic loss. Although the result of Average Returns converges quickly and the pendulum works fine, the loss never converges even with over 200,000 iterations. To deal with this problem, batch size is set to 32, 64, 100 and target network update period is set to 5, 10, 100, 1000, 10000 respectively according to \cite{mnih2015human}. However, except for longer running time and several crashes of Colab, nothing helps with converging and decreasing loss values. After reading lots of blogs and papers, loss function is replaced with Mean Absolute Error(MAE) ultimately. It gives relative better result than other two functions in terms of converging training loss, which might because of its robustness to outliers.

\begin{figure}[H]
    \centering
    \includegraphics[scale=0.3]{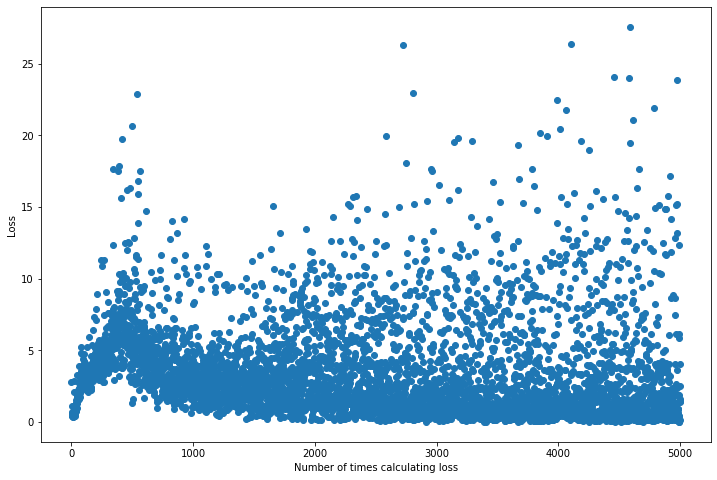}
    \caption{Training Loss Every 20 Iterations}
    \label{fig:4}
\end{figure}

\begin{figure}[H]
    \centering
    \includegraphics[scale=0.3]{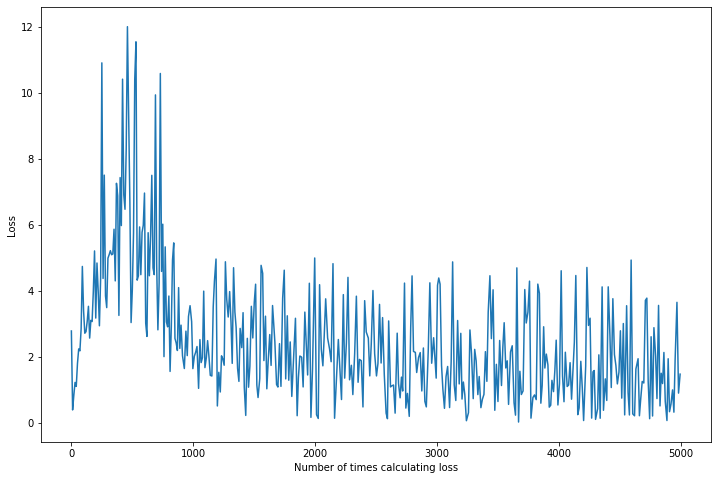}
    \caption{Training Loss Every 20 Iterations After Filtering}
    \label{fig:5}
\end{figure}

Figure \ref{fig:4} shows the loss value every 20 iterations using MAE. Clearly, there is a small increase at the beginning, then the loss decreases and approximately converges with some ``outliers'' showing learning process. For clear view, after initial unstable steps, loss values less than 5, which weights around $89\%$ of data, were kept and plotted again in Figure \ref{fig:5}. Now obviously, loss stables as learning further and eventually converges around 2. 

\begin{figure}[H]
\centering
\begin{subfigure}{0.15\textwidth}
  \centering
  \includegraphics[width=\linewidth]{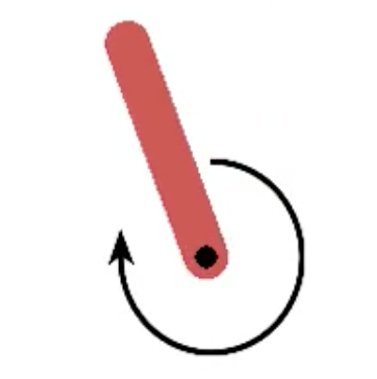}
  \caption{MSE}
  \label{fig:6a}
\end{subfigure}%
\begin{subfigure}{0.15\textwidth}
  \centering
  \includegraphics[width=\linewidth]{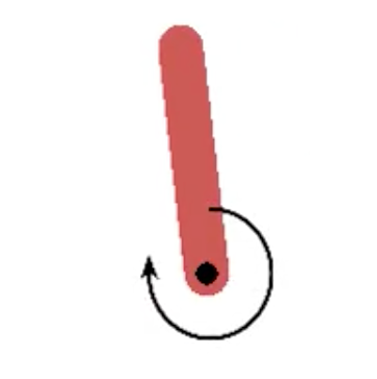}
  \caption{Huber}
  \label{fig:6b}
\end{subfigure}%
\begin{subfigure}{0.15\textwidth}
  \centering
  \includegraphics[width=\linewidth]{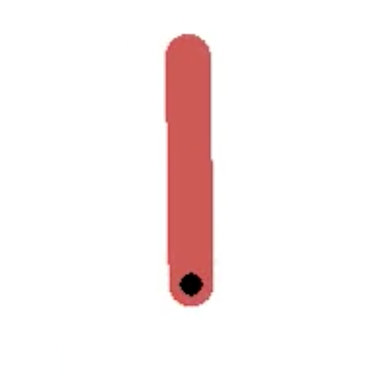}
  \caption{MAE}
  \label{fig:6c}
\end{subfigure}%
\vspace{5pt}
\caption{Pendulum Position with Different Loss Function} 
\label{fig:6}
\end{figure}

MAE also works better in keeping pendulum upright. Pendulum with MSE (Figure \ref{fig:6a}) and Huber (Figure \ref{fig:6b}) can stay above horizontal line but always stopped slightly away from vertical while using MAE can keep the pendulum exactly upright (Figure \ref{fig:6c}).

\vspace{5pt}
\section{Conclusion}
\label{4}
Using DDPG algorithm, it becomes possible to solve challenging problems over continuous action spaces in fewer steps. However, DDPG requires a large number of episodes to find its optimal solution due to the property of model-free reinforcement learning approaches. The availability of TF-Agents helps us easily conduct RL experiments. Based on our results, the pendulum can learn well after being trained with DDPG agent and its average return quickly converges. In future research, we can apply this agent to more complicated games like Bipedal-Walker or Alien-ram. Moreover, creating new loss function might be useful for tracking a stable loss and further improving the learned pendulum.

\bibliographystyle{IEEEtran}
\bibliography{refs}

\end{document}